\documentclass[10pt,twocolumn,letterpaper]{article}

\usepackage{cvpr}
\usepackage{times}
\usepackage{epsfig}
\usepackage{graphicx}
\usepackage{amsmath}
\usepackage{amssymb}

\usepackage{multirow}
\usepackage{subcaption}
\usepackage{algorithm}
\usepackage{authblk}

\usepackage{algpseudocode}

\usepackage[breaklinks=true,bookmarks=false]{hyperref}

\graphicspath{ {img/} }
\newtheorem{mytheo}{Proposition}

\cvprfinalcopy 

\def\cvprPaperID{3809} 

\ifcvprfinal\fi
\begin{document}
\title{Learning Time/Memory-Efficient Deep Architectures with Budgeted Super Networks}


\author[*]{Tom V\'eniat}
\author[*,\dag]{Ludovic Denoyer}
\affil[*]{Sorbonne Universit\'e, LIP6, F-75005, Paris, France}
\affil[ \dag]{Criteo Research}
\affil[ ]{\texttt{\{tom.veniat, ludovic.denoyer\}@lip6.fr}}

\maketitle
\thispagestyle{empty}

\begin{abstract}
 We propose to focus on the problem of discovering neural network architectures efficient in terms of both prediction quality and cost. For instance, our approach is able to solve the following tasks: learn a neural network able to predict well in less than 100 milliseconds or learn an efficient model that fits in a 50 Mb memory. Our contribution is a novel family of models called Budgeted Super Networks (BSN). They are learned using gradient descent techniques applied on a budgeted learning objective function which integrates a maximum authorized cost, while making no assumption on the nature of this cost. We present a set of experiments on computer vision problems and analyze the ability of our technique to deal with three different costs: the computation cost, the memory consumption cost and a distributed computation cost. We particularly show that our model can discover neural network architectures that have a better accuracy than the ResNet and Convolutional Neural Fabrics architectures on CIFAR-10 and CIFAR-100, at a lower cost. 

\end{abstract}

\section{Introduction}

In the Deep Learning community, finding the best Neural Network architecture for a given task is a key problem that is mainly addressed \textit{by hand} or using validation techniques. For instance, in computer vision, this has lead to particularly well-known models like GoogleNet \cite{DBLP:journals/corr/SzegedyLJSRAEVR14} or ResNet \cite{DBLP:journals/corr/HeZRS15}. More recently, there is a surge of interest in developing techniques able to automatically discover efficient neural network architectures. Different algorithms have been proposed including evolutionary methods \cite{DBLP:conf/gecco/StanleyM02a, DBLP:journals/corr/MiikkulainenLMR17, DBLP:journals/corr/RealMSSSLK17} or reinforcement learning-based approaches \cite{DBLP:journals/corr/ZophL16}. But in all cases, this selection is usually based solely on a final predictive performance of the model such as the accuracy. 

When facing real-world problems, this predictive performance is not the only measure that matters. Indeed, learning a very good predictive model with the help of a cluster of GPUs might lead to a neural network that can be incompatible with low-resource mobile devices. Another example concerns distributed models in which one part of the computation is made \textit{in the cloud} and the other part is made \textit{on the device}. In such situations, an efficient architecture would have to predict accurately while minimizing the amount of exchanged messages between the cloud and the device. One important research direction is thus to propose models that can learn to take into account the inference cost in addition to the quality of the prediction. 

We formulate this issue as a problem of automatically learning a neural network architecture under budget constraints. To tackle this problem, we propose a \textit{budgeted learning} approach that integrates a maximum cost directly in the learning objective function. The main originality of our approach with respect to state-of-the-art is the fact that it can be used with any type of costs, existing methods being usually specific to particular constraints like inference speed or memory consumption -- see Section \ref{sec_related_work} for a review of state-of-the-art. In our case, we investigate the ability of our method to deal with three different costs: (i) the \textit{computation cost} reflecting the inference speed of the resulting model, (ii) \textit{the memory consumption cost} that measures the final size of the model, and the (iii) \textit{distributed computation cost} that measures the inference speed when computations are distributed over multiple machines or processors.

\begin{figure*}[ht]
\begin{center}
\begin{subfigure}[t]{.49\textwidth} 
	\centering    
    \includegraphics[width=0.6\linewidth]{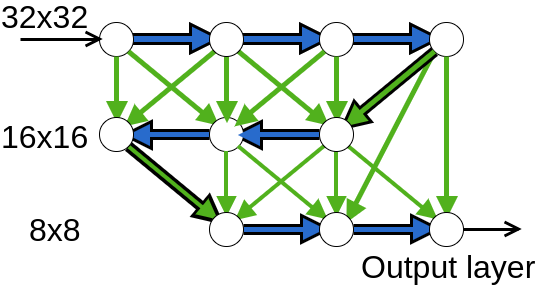}
    \caption{\textbf{ResNet Fabric}: The ResNet Fabric is a super network that includes the ResNet model as a particular sub-graph. Each row corresponds to a particular size and number of feature maps. Each edge represents a simple \textit{building block} (as described in\cite{DBLP:journals/corr/HeZRS15}) i.e two stacked convolution layers + a shortcut connection. We use projection shortcuts (with 1x1 convolutions) for all connections going across different feature map sizes (green edges). Note that here, the subgraph corresponding to the bold edges is a ResNet-20. By increasing the width of the ResNet Fabric, we can include different variants of ResNets (from ResNet-20 with width 3 up to Resnet-110 with a width of 18). }
	\label{fig1a}
\end{subfigure}
\hfill
\begin{subfigure}[t]{.49\textwidth}
	\centering
	\includegraphics[width=0.85\linewidth]{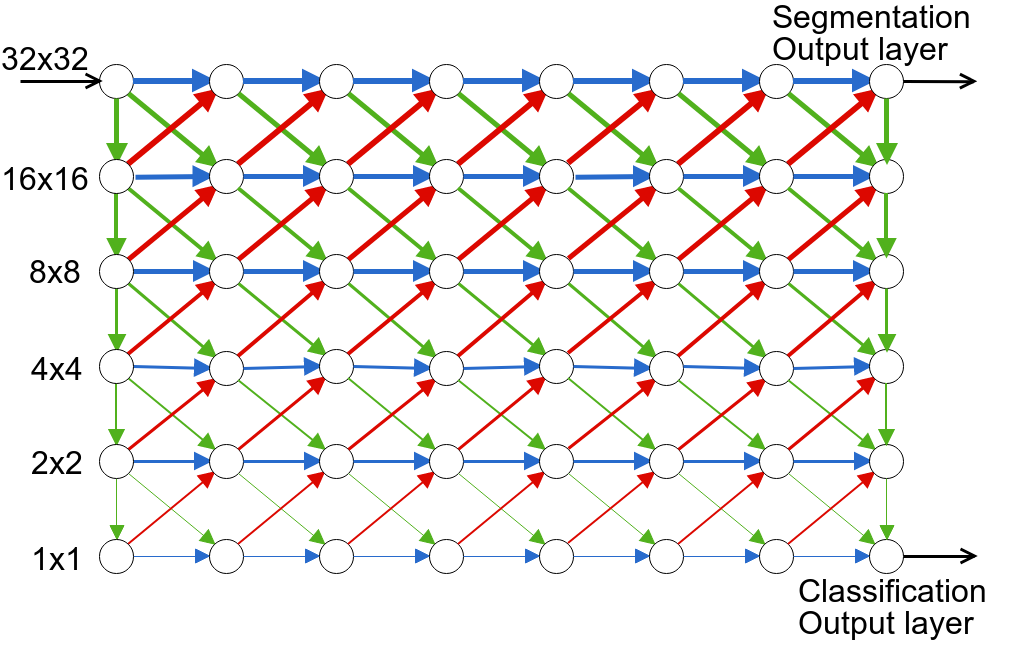}         
	\caption{\textbf{Convolutional Neural Fabrics} \cite{DBLP:journals/corr/SaxenaV16}: Each row corresponds to a particular resolution of feature maps. The number of features map is constant across the whole network. Each edge represents a convolution layer. The color of an edge represents the difference between input and output maps resolutions. Blue edges keep the same resolution, green edges decrease the resolution (stride $>$ 1) and red edges increase the resolution (upsampling). Feature maps are aggregated (by addition) at each node before being sent to the next layers. }
	\label{fig1b}
\end{subfigure}

\end{center}
   \caption{This figure illustrates the two Super Networks on top of which cost-constrained architectures will be discovered. The ResNet Fabric is a generalization of ResNets\cite{DBLP:journals/corr/HeZRS15}, while CNF has been proposed in \cite{DBLP:journals/corr/SaxenaV16}. In both cases, our objective is to discover architectures that are efficient in both prediction quality and cost, by sampling edges over these S-networks.}
\label{fig:SuperNetworks}
\label{SuperNetworks_imgs}
\end{figure*}

Our model called \textit{Budgeted Super Network} (BSN) is based on the following principles: (i) the user provides a (big) \textit{Super Network} (see Section \ref{secsn}) defining a large set of possible final network architectures as well as a maximum authorized cost. (ii) Since finding the best architecture that satisfies the cost constraint is an intractable combinatorial problem (Section \ref{s3}), we relax this optimization problem and propose a stochastic model (called \textit{Stochastic Super Networks} -- Section \ref{s4}) that can be optimized using policy gradient-inspired methods (Section \ref{s5}). We show that the optimal solution of this stochastic problem corresponds to the optimal constrained network architecture (Proposition \ref{prop1}) validating our approach. At last, we evaluate this model on various computer vision tasks. We particularly show that, by taking inspiration from the \textit{Residual Networks} (ResNet) \cite{DBLP:journals/corr/HeZRS15} and \textit{Convolutional Neural Fabrics} (CNF) \cite{DBLP:journals/corr/SaxenaV16}, our model is able to discover new neural network architectures that outperform these baselines at a lower computation/memory/distributed cost (Section \ref{section_experiments}) on CIFAR-10 and CIFAR-100. The related work is presented in Section \ref{sec_related_work}.


\section{Super Networks}
\label{secsn}
We consider the classical supervised learning problem defined by an input space $\mathcal{X}$ and an output space $\mathcal{Y}$. In the following, input and output spaces correspond to multi-dimensional real-valued spaces. The training set is denoted as $\mathcal{D}=\{ (x^1,y^1), ..., (x^\ell,y^\ell) \}$ where $x^i \in \mathcal{X}$, $y^i \in \mathcal{Y}$ and $\ell$ is the number of supervised examples.  At last, we consider a model $f:  \mathcal{X} \rightarrow \mathcal{Y}$ that predicts an output given a particular input. 

We first describe a family of models called \textit{Super Networks} (S-networks)\footnote{The name \textit{Super Network} comes from \cite{DBLP:journals/corr/FernandoBBZHRPW17} which presents an architecture close to ours for a completely different purpose. } since our contribution presented in Section \ref{section_super_net} will be a stochastic extension of this model. Note that the principle of Super Networks is not new and similar ideas have been already proposed in the literature under different names, e.g Deep Sequential Neural Networks \cite{DBLP:journals/corr/DenoyerG14}, Neural Fabrics \cite{DBLP:journals/corr/SaxenaV16}, or even PathNet \cite{DBLP:journals/corr/FernandoBBZHRPW17}.

A Super Network is composed of a set of layers connected together in a direct acyclic graph (DAG) structure. Each edge is a (small) neural network, the S-Network corresponds to a particular combination of these neural networks and defines a computation graph. Examples of S-networks are given in Figure \ref{SuperNetworks_imgs}. More formally, let us denote $l_1,....,l_N$ a set of layers, $N$ being the number of layers, such that each layer $l_i$ is associated with a particular representation space $\mathcal{X}_i$ which is a  multi-dimensional real-valued space. $l_1$ will be the \textit{input layer} while $l_N$ will be the \textit{output layer}. We also consider a set of (differentiable) functions $f_{i,j}$ associated to each possible pair of layers such that $f_{i,j}: \mathcal{X}_i\rightarrow \mathcal{X}_j$. Each function $f_{i,j}$ will be referred as a \textit{module} in the following: it takes data from $\mathcal{X}_i$ as inputs and transforms these data to  $\mathcal{X}_j$. Note that each $f_{i,j}$ will make disk/memory/network operations having consequences on the inference speed of the S-network. Each $f_{i,j}$ module is associated with parameters in $\theta$, $\theta$ being implicit in the notation for sake of clarity.

On top of this structure, a particular architecture $E=\{e_{i,j}\}_{(i,j) \in [1;N]^2}$ is a binary adjacency matrix over the $N$ layers such that $E$ defines a DAG with a single source node $l_1$ and a single sink node $l_N$. Different matrices $E$ will thus correspond to different super network architectures. A S-network will be denoted $(E,\theta)$ in the following, $\theta$ being the parameters of the different \textit{modules}, and $E$ being the architecture of the super network.

\paragraph{Predicting with S-networks:} The computation of the output $f(x,E,\theta)$ given an input $x$ and a S-network $(E,\theta)$ is made through a classic forward algorithm, the main idea being that the output of modules $f_{i,j}$ and $f_{k,j}$ leading to the same layer $l_j$ will be added in order to compute the value of $l_j$. Let us denote $l_i(x,E,\theta)$ the value of layer $l_i$ for input $x$, the computation is recursively defined as: 
\begin{equation}
\begin{aligned}
\text{Input:} l_1(x,E,\theta) &= x \\
\text{Layer Computation: } l_i(x,E,\theta) &= \sum\limits_k e_{k,i} f_{k,i}(l_k(x,E,\theta))
\end{aligned}
\end{equation}
 In this configuration, learning of $\theta$ can be made using classical back-propagation and gradient-descent techniques.

\section{Learning Cost-constrained architectures}
\label{section_super_net}

Our main idea is the following: we now consider that the structure $E$ of the S-network $(E,\theta)$ describes not a single neural network architecture but a set of possible architectures. Indeed, each sub-graph of $E$ (subset of edges)  corresponds itself to a S-network and will be denoted $H \odot E$, where $H$ corresponds to a binary matrix used as a mask to select the edges in $E$ and $\odot$ is the Hadamard product. Our objective will thus be to identify the best matrix $H$  such that the corresponding S-network $(H \odot E,\theta)$ will be a network efficient in terms of both predictive quality and computation/memory/... cost. 

The next sections are organized as follows: (i) First, we formalize this problem as a combinatorial problem where one wants to discover the best matrix $H$ in the set of all possible binary matrices of size $N \times N$. Since this optimization problem is intractable, we propose a new family of models called \textit{Stochastic Super Networks} where $E$ is sampled following a parametrized distribution $\Gamma$ before each prediction. We then show that the resulting budgeted learning problem is continuous and that its solution corresponds to the optimal solution of the initial budgeted problem (Proposition 1). We then propose a practical learning algorithm to learn $\Gamma$ and $\theta$ simultaneously using gradient descent techniques.

\subsection{Budgeted Architectures Learning} 
\label{s3}
Let us consider $H$ a binary matrix of size $N \times N$. Let us denote $C(H \odot E) \in \mathbb{R}^+$ the cost\footnote{Note that we consider that the cost only depends on the network architecture. The model could easily be extended to costs that depend on the input $x$ to process, or to stochastic costs -- see appendix} associated to the computation of the S-Network $(H \odot E,\theta)$. Let us also define $\mathbf{C}$ the maximum cost the user would allow. For instance, when solving the problem of \textit{learning a model with a computation time lower than 200 ms} then $\mathbf{C}$ is equal to $200 ms$. We aim at solving the following soft constrained budgeted learning problem:

\begin{multline} \label{objective}
H^*,\theta^* = \arg \min\limits_{H,\theta} \frac{1}{\ell} \sum\limits_i \Delta(f(x^i,H \odot E, \theta),y^i)
\\ + \, \lambda \max(0,C(H \odot E)-\mathbf{C})
\end{multline}
where $\lambda$ corresponds to the importance of the cost penalty. Note that the evaluated cost is specific to the particular infrastructure on which the model is ran. For instance, if $\mathbf{C}$ is the cost in milliseconds, the value of $C(H \odot E)$ will not be the same depending on the device on which the model is used. Note that the only required property of $C(H \odot E)$ is that this cost can be measured during training.

Finding a solution to this learning problem is not trivial since it involves the computation of all possible architectures which is prohibitive ($\mathcal{O}(2^N)$ in the worst case). We explain in the next section how this problem can be solved using \textit{Stochastic Super Networks}. 

\begin{figure}[t]
\centering
\includegraphics[width=1.1\linewidth]{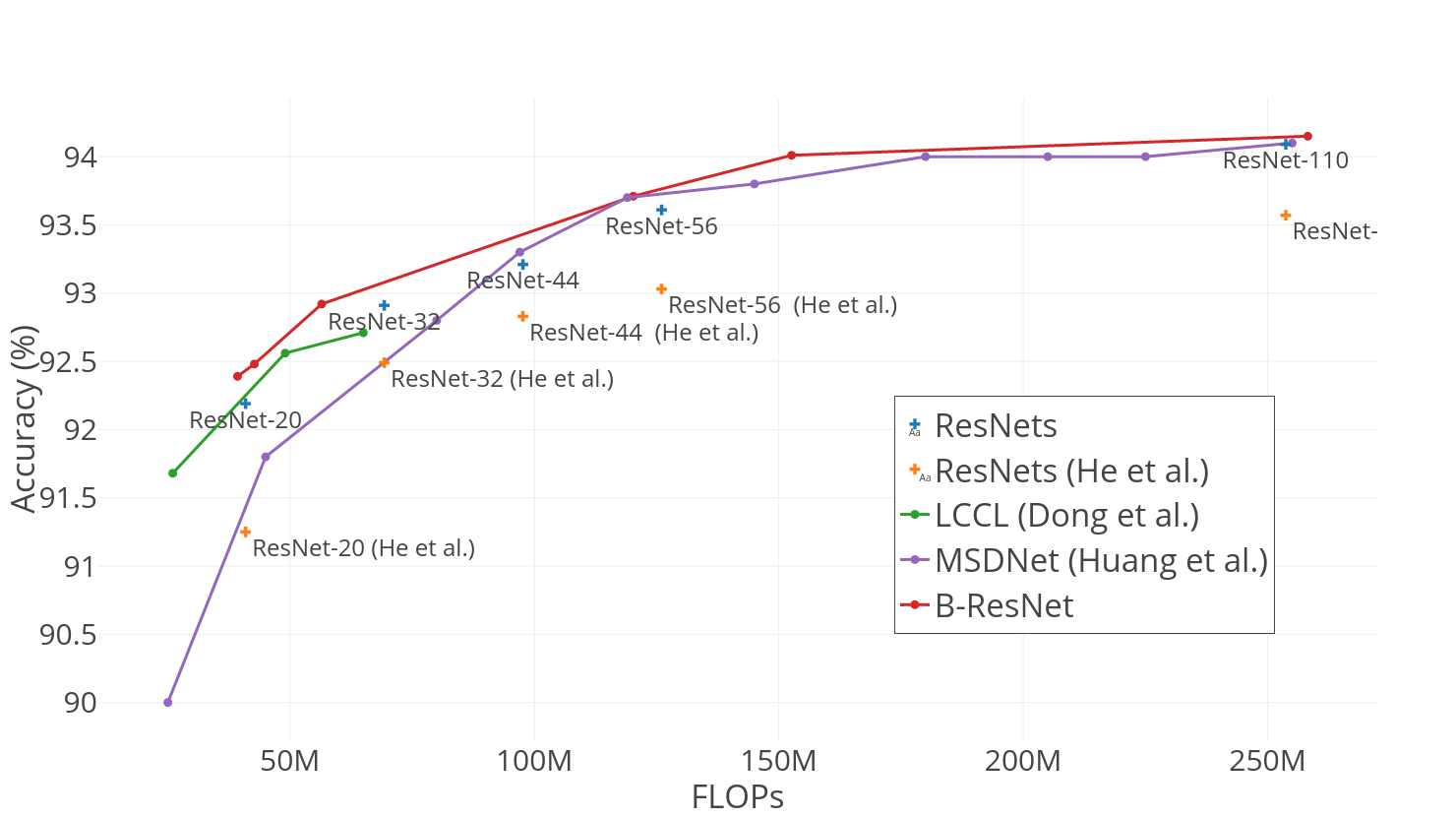} 
\caption{Accuracy/Time trade-off using B-ResNet on CIFAR-10.}
\label{plot_cif10_flop_rnf}
\end{figure}

\subsection{Stochastic Super Networks}
\label{secion_ssn}
\label{s4}



Now, given a particular architecture $E$, we consider the following stochastic model -- called \textbf{Stochastic Super Network} (SS-network) -- that computes a prediction in two steps: 
\begin{enumerate}
\item A binary matrix $H$ is sampled based on a distribution with parameters $\Gamma$. This operation is denoted $H \sim \Gamma$ 
\item The final prediction is made using the associated sub-graph i.e. by computing $f(x,H \odot E,\theta)$. 
\end{enumerate}
A SS-network is thus defined by a triplet $(E,\Gamma,\theta)$, where both $\Gamma$ and $\theta$ are learnable parameters. 

We can rewrite the budgeted learning objective of Equation \ref{objective} as:
\begin{multline} \label{stochobjective}
 \Gamma^* , \theta^* = \arg \min\limits_{\Gamma,\theta} \frac{1}{\ell} \sum\limits_i \mathbb{E}_{H \sim \Gamma}\left[ \Delta(f(x^i,H \odot E, \theta),y^i) \right. \\+ \left. \lambda \max (0,C(H \odot E)-\mathbf{C}) \vphantom{\Delta}\right]
\end{multline}

\begin{mytheo}
\label{prop1} (proof in Appendix) 
When the solution of Equation \ref{stochobjective} is reached, then the models sampled following $(\Gamma^*)$ and using parameters $\theta^*$ are optimal solution of the problem of Equation \ref{objective}. 
\end{mytheo}

Said otherwise, solving the stochastic problem will provide a model that has a good predictive performance under the given cost constraint. 

\paragraph{Edge Sampling: } In order to avoid inconsistent architectures where the input and the output layers are not connected, we sample $H$ using the following procedure: For each layer $l_i$ visited in the topological order of $E$ (from the first layer to the last one) and for all $k<i$:  If $l_k$ is connected to the input layer $l_1$ based on the previously sampled edges, then $h_{k,i}$ is sampled following a Bernoulli distribution with probability\footnote{Note that $\gamma_{k,i}$ is obtained by applying a \textit{logistic} function over a continuous parameter value, but this is made implicit in our notations.} $\gamma_{k,i}$. In the other cases, $h_{k,i}=0$. 

\subsection{Learning Algorithm}
\label{s5}

We consider the generic situation where the cost-function $C(.)$ is unknown and can be observed at the end of the computation of the model over an input $x$. Note that this case also includes stochastic costs where $C$ is a random variable, caused by some network latency during distributed computation for example. We now describe  the case where $C$ is deterministic, see appendix  for its stochastic extension.

Let us denote $\mathcal{D}(x,y,\theta,E,H)$ the quality of the S-Network $(H \odot E,\theta)$ on a given training pair $(x,y)$: 
\begin{multline} \label{D_def}
\mathcal{D}(x,y,\theta,E,H)=\Delta(f(x,H \odot E,\theta),y) \\+ \lambda \max(0, C(H \odot E) - \mathbf{C})
\end{multline}
We propose to use a policy gradient inspired algorithm as in \cite{DBLP:journals/corr/DenoyerG14,DBLP:journals/corr/BengioBPP15} to learn $\theta$ and $\Gamma$. Let us denote $\mathcal{L}(x,y,E,\Gamma,\theta)$ the expectation of $\mathcal{D}$ over the possible sampled matrices $H$:
\begin{equation}
\mathcal{L}(x,y,E,\Gamma,\theta) = \mathbb{E}_{H \sim \Gamma} \mathcal{D}(x,y,\theta,E,H)
\end{equation}

The gradient of $\mathcal{L}$ can be written as\footnote{details provided in appendix}:
\begin{multline}
\nabla_{\theta,\Gamma} \mathcal{L}(x,y,E,\Gamma,\theta) \\ = \sum\limits_H P(H|\Gamma) \left[(\nabla_{\theta,\Gamma}  \log P(H|\Gamma)) \mathcal{D}(x,y,\theta,E,H) \right] \\+ \sum\limits_H P(H|\Gamma) \left[ \nabla_{\theta,\Gamma} \Delta(f(x,H \odot E,\theta),y) \right]
\end{multline}
The first term corresponds to the gradient over the log-probability of the sampled structure $H$ while the second term is the gradient of the prediction loss given the sampled structure $H \odot E$.

Learning can be made using back-propagation and stochastic-gradient descent algorithms as it is made in Deep Reinforcement Learning models. Note that in practice, in order to reduce the variance of the estimator, the update is made following:
\begin{multline}
\nabla_{\theta,\Gamma} \mathcal{L}(x,y,E,\Gamma,\theta) \\ \approx (\nabla_{\theta,\Gamma} \log P(H|\Gamma)) (\mathcal{D}(x,y,\theta,E,H)-\tilde{\mathcal{D}} )  \\+ \nabla_{\theta,\Gamma} \Delta(f(x,H \odot E,\theta),y)
\end{multline}
where $H$ is sampled following $\Gamma$, and where $\tilde{\mathcal{D}}$ is the average value of $\mathcal{D}(x,y,\theta,E,H)$ computed on a batch of learning examples.

\section{Experiments}\label{section_experiments}

\subsection{Implementation}\label{implem}

We study two particular S-Network architectures:

\textbf{ResNet Fabric} (Figure \ref{fig1a}) which is used for \textbf{image classification}. This S-Network is inspired by the ResNet\cite{DBLP:journals/corr/HeZRS15} architecture on which extra modules (i.e. edges) have been added. The underlying idea is that a particular sub-graph of the ResNet Fabric corresponds exactly to a ResNet model. We thus aims at testing the ability of our approach to discover ResNet-inspired efficient architectures, or at least to converge to a ResNet model that is known to be efficient.

\textbf{Convolutional Neural Fabrics} (CNF) which has been proposed in \cite{DBLP:journals/corr/SaxenaV16} (Figure \ref{fig1b}). It is a generic architecture that can be used for both \textbf{image classification} and \textbf{image segmentation}. The layers of the CNF super network are organized in a $W \times H$ matrix. We always use $W=8$ when running our budgeted algorithm. Different values of $W$ (as in \cite{DBLP:journals/corr/SaxenaV16}) are used as baselines. 

Image classification has been tested on CIFAR-10 and CIFAR-100 \cite{Krizhevsky09learningmultiple} while the image segmentation has been performed on the Part Label dataset \cite{GLOC_CVPR13}. 

For these two architectures denoted \textit{B-ResNet} and \textit{B-CNF}, we consider three different costs functions: the first one is the (i) \textit{computation cost} computed as the number of operations \footnote{The number of Mult-Add operations required to fully evaluate a network.} made by the S-Network as used in \cite{DBLP:journals/corr/DongHYY17} or \cite{DBLP:journals/corr/HuangCLWMW17}. Note that this cost is highly correlated with the execution time\footnote{Expressing constraint directly in \textit{milliseconds} has been also investigated, with results similar to the ones obtain using the computation cost, and are not presented here.}. The second one is the \textit{memory consumption cost}, measured as the number of parameters of the resulting models. At last, the third cost (iii) is the \textit{distributed computation cost} which is detailed in Section \ref{parallel_section} and corresponds to the ability of a particular model to be efficiently computed over a distributed environment. 

\subsection{Experimental Protocol and Baselines}

Each model is trained with various values for the objective cost $\mathbf{C}$. For the image classification problem, since we directly compare to ResNet, we select values of $\mathbf{C}$ that corresponds to the cost of the ResNet-20/32/44/56/110 architectures. This allows us to compare the performance of our method at the same cost level as the ResNet variants. When dealing with the B-CNF model, we select $\mathbf{C}$ to be the cost of different versions of the CNF model, having different width W. The height H being fixed by the resolution of the input image.

For each experiment, multiple versions of the different models are evaluated over the validation set during learning. Since our evaluation now involves both a cost and an accuracy, we select the best models using the pareto front on the cost/accuracy curve on the validation set. The reported performance are then obtained by evaluating these selected models over the test set. The detailed procedure of the hyper-parameters and model selection are given in Appendix and the source code of our implementation is open source\footnote{\url{https://github.com/TomVeniat/bsn}}. The learning is done using a classical stochastic gradient-descent algorithm for all parameters with learning rate decay, momentum of 0.9 and weight decay of $10^{-4}$ for $\theta$.

\begin{table}[h]
\centering
\begin{tabular}{|c||cc|}
\hline
Model                         & FLOPs (millions)   & Accuracy                    \\ \hline \hline
\multicolumn{2}{|c|}{ResNet \cite{DBLP:journals/corr/HeZRS15}}                       & our/\textit{original} \\ \hline
ResNet-110                    & 253.70              & 94.09/\textit{93.57}                 \\
ResNet-56                     & 126.01             & 93.61/\textit{93.03}                 \\
ResNet-44                     & 97.64              & 93.21/\textit{92.83}                 \\
ResNet-32                     & 69.27              & 92.91/\textit{92.49}                 \\
ResNet-20                     & 40.90               & 92.19/\textit{91.25}                 \\ \hline \hline
\multicolumn{3}{|c|}{Low Cost Collaborative Layer \cite{DBLP:journals/corr/DongHYY17}}                               \\ \hline
LCCL (ResNet-110)             & 166            & 93.44                       \\
LCCL (ResNet-44)              & 65                 & 92.71                       \\
LCCL (ResNet-32)              & 49                 & 92.56                       \\
LCCL (ResNet-20)              & 26                 & 91.68                       \\ \hline \hline
\multicolumn{3}{|c|}{Multi Scale DenseNet \cite{DBLP:journals/corr/HuangCLWMW17} (values read on plot)}                         \\ \hline
\multirow{10}{*}{MSDNet}      & $\approx 255$                & 94.1			\\
                              & $\approx 225$                & 94.0			\\
                              & $\approx 205$                & 94.0           \\
                              & $\approx 180$                & 94.0           \\
                              & $\approx 145$                & 93.8         \\
                              & $\approx 119$                & 93.7         \\
                              & $\approx 97 $                & 93.3         \\
                              & $\approx 80 $                & 92.8         \\
                              & $\approx 45 $                & 91.8         \\
                              & $\approx 25 $                & 90.0           \\ \hline \hline
\multicolumn{3}{|c|}{Budgeted ResNet}                                       \\ \hline
\multirow{7}{*}{B-ResNet}     & 407.51             & 94.29                       \\
                              & 258.20              & 94.15                       \\
                              & 152.60              & 94.01                       \\
                              & 120.20              & 93.71                       \\
                              & 56.47              & 92.92                       \\
                              & 42.69              & 92.48                       \\
                              & 39.25              & 92.39                       \\ \hline \hline
\multicolumn{2}{|c|}{Convolutional Neural Fabrics \cite{DBLP:journals/corr/SaxenaV16}} & our/\textit{original} \\ \hline
CNF W=8                       & 2,219.00           & 94.83/\textit{90.58}               \\
CNF W=4                       & 1,010.00           & 93.75/\textit{87.91}               \\
CNF W=2                       & 406.00             & 92.54/\textit{86.21}               \\
CNF W=1                       & 54.00              & 89.91                       \\ \hline \hline
\multicolumn{3}{|c|}{Budgeted CNF}                                               \\ \hline
\multirow{5}{*}{B-CNF}        & 2,150.00           & 94.92                       \\
                              & 1,407.00           & 94.85                       \\
                              & 1,144.00           & 94.69                       \\
                              & 103.00             & 93.14                       \\
                              & 85.00              & 92.17                       \\ \hline
\end{tabular}
\caption{Accuracy/speed trade-off on CIFAR-10 using B-ResNet and B-CNF. Values reported as \textit{our} corresponds to results we obtained when training a reproduction of the models, \textit{original} corresponds to values from the original article.}
\label{cif10_allmodels_flop}
\end{table}

For each experiment, we give the performance of both reference models (ResNet \cite{DBLP:journals/corr/HeZRS15} and CNF \cite{DBLP:journals/corr/SaxenaV16}), and of related existing models
i.e Low Cost Collaborative Layer (LCCL)\cite{DBLP:journals/corr/DongHYY17} and MSDNet \cite{DBLP:journals/corr/HuangCLWMW17} (under the anytime classification settings). Note that the baselines methods have been designed to reduce exclusively the \textit{computation cost}, while our technique is able to deal with any type of cost. We provide the performance of our budgeted version of ResNet (B-ResNet) and of our budgeted version of CNF (B-CNF). Note that, for a fair comparison, we present the published results of ResNet and CNF, but also the ones that we have obtained by training these models by ourselves, \textbf{our results being of better quality} than the previously published performance.

\subsubsection{Experimental results}\label{section_classif}


\begin{figure*}[ht]
\begin{center}

\begin{subfigure}[t]{.3\textwidth}
	\centering
	\includegraphics[width=\linewidth]{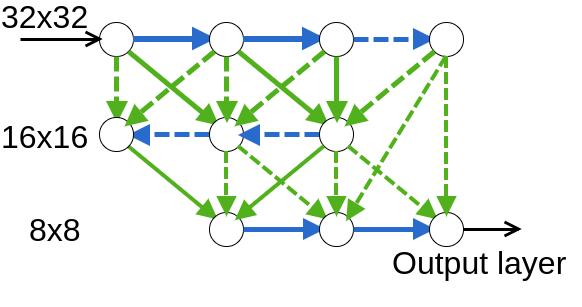}         
	\caption{B-ResNet}
	\label{arch_resnet}
\end{subfigure}
\begin{subfigure}[t]{.3\textwidth} 
	\centering
    \includegraphics[width=\linewidth]{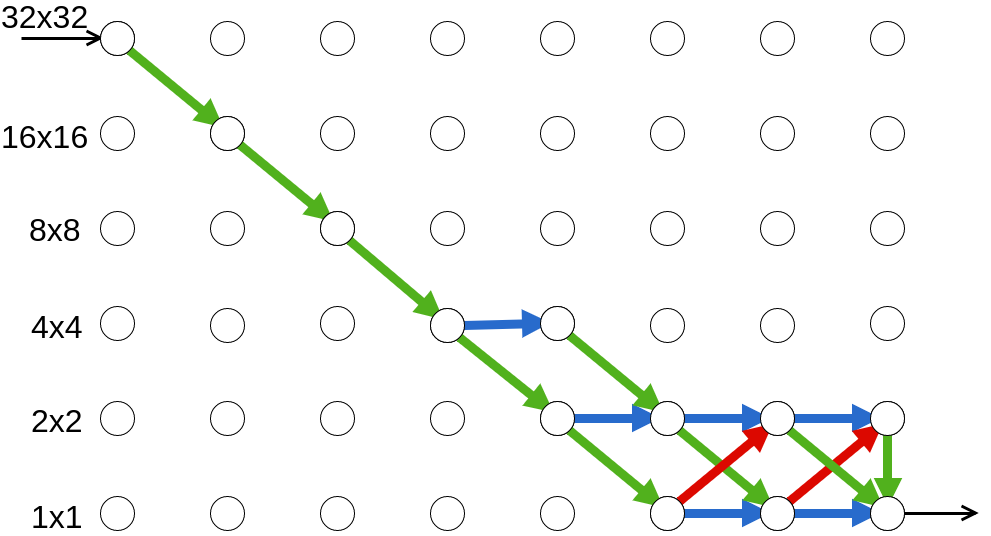}
    \caption{B-CNF \& computation cost}
	\label{arch_cnf_flop}
\end{subfigure}
\begin{subfigure}[t]{.3\textwidth}
	\centering
	\includegraphics[width=\linewidth]{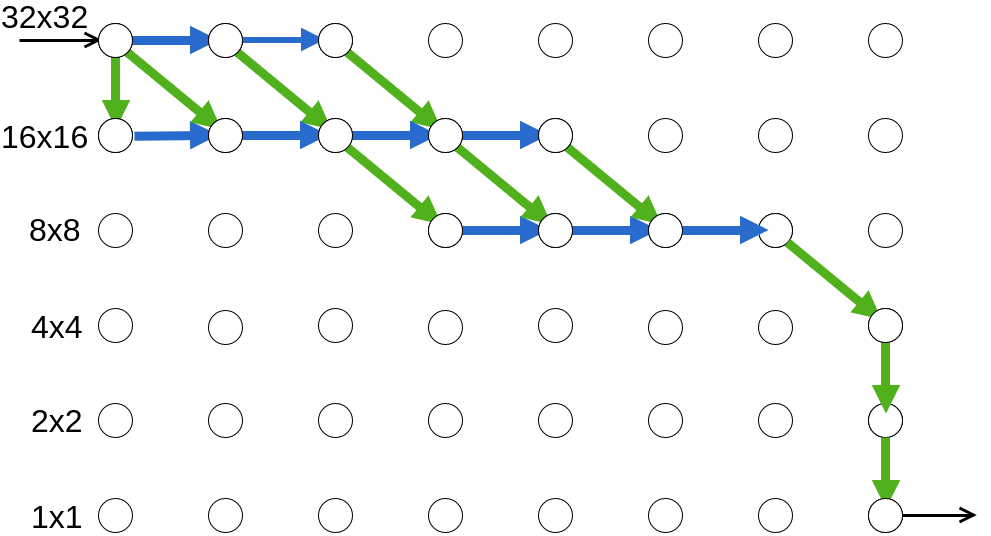}  
	\caption{B-CNF \& memory consumption cost}
	\label{arch_cnf_param}
\end{subfigure}

\end{center}
   \caption{Discovered architectures: \textbf{(Left)} is a low \textit{computation cost} B-ResNet where dashed edges correspond to connections in which the two convolution layers have been removed (only shortcut or projection connections are kept). \textbf{(Center)} is a low \textit{computation cost} B-CNF where high-resolution operations have been removed. \textbf{(Right)} is a low \textit{memory consumption cost} B-CNF: the algorithm has mostly kept all high resolution convolutions since they allow fine-grained feature maps and have the same number of parameters than lower-resolution convolutions. It is interesting to note that our algorithm, constrained with two different costs, automatically learned two different efficient architectures.} 
\label{fig:DiscoveredNetworks}
\end{figure*}

\begin{table}[ht]
\centering
\small{
\begin{tabular}{|c||cc|}
\hline
Model                 & FLOPs (millions) & Accuracy (\%) \\ \hline \hline
ResNet-110            & 253.7            & 71.85         \\
ResNet-56             & 126              & 70.57         \\
ResNet-44             & 97.64            & 70.28         \\
ResNet-32             & 69.27            & 69.28         \\
ResNet-20             & 40.9             & 67.14         \\ \hline
\multirow{7}{*}{MSDNet \cite{DBLP:journals/corr/HuangCLWMW17}} & 215              & 76            \\
 		& 180              & 75            \\
		& 150              & 74            \\
 		& 109              & 72.5          \\
		& 80               & 71            \\
		& 45               & 67.5          \\
		& 15               & 62.5          \\ \hline
\multirow{7}{*}{B-ResNet}	& 349.5            & 73.28         \\
							& 115.09           & 71.46         \\
							& 69.84            & 70.27         \\
			          		& 64.96            & 70.12         \\
			          		& 46.29            & 69.02         \\
			          		& 39.22            & 68.45         \\ \hline
\end{tabular}
}
\caption{Accuracy/speed trade-off on Cifar-100 using ResNet Fabrics.}
\label{cif100_resnetfab_flop}
\end{table}

\paragraph{Reducing the computation cost:} Figure \ref{plot_cif10_flop_rnf} and Table \ref{cif10_allmodels_flop} show the performance of different models over CIFAR-10. Each point corresponds to a model evaluated both in term of accuracy and \textit{computation cost}. When considering the B-ResNet model, and by fixing the value of $\mathbf{C}$ to the computation cost of the different ResNet architectures, we obtain budgeted models that have approximatively the same costs than the ResNets, but with a higher accuracy. For example, ResNet-20 obtains an accuracy of 92.19\% at a cost of $40.9 \times 10^6$ flop, while B-ResNet is able to discover an architecture with 92.39\% accuracy at a slightly lower cost ($39.25 \times 10^6$ flop). Moreover, the B-ResNet model also outperforms existing approaches like MSDNet or LCCL, particularly when the \textit{computation cost} is low i.e for architectures that can be computed at a high speed. When comparing CNF to B-CNF, one can see that our approach is able to considerably reduce the computation cost while keeping a high accuracy. For example, one of our learned models obtained an accuracy of 93.14\% with a cost of $103 \times 10^6$ flop while CNF has an accuracy of 92.54\% for a cost of $406 \times 10^6$ flop. Note that the same observations can be drawn for CIFAR-100 (Table \ref{cif100_resnetfab_flop}). 

Figure \ref{arch_resnet} and \ref{arch_cnf_flop} illustrate two architectures discovered by B-ResNet and B-CNF with a low \textit{computation cost}. One can see that B-ResNet has converged to an architecture which is a little bit different than the standard ResNet architecture, explaining why its accuracy is better. On the CNF side, our technique is able to extract a model that has a minimum of high-resolution convolutions operations, resulting in a high speedup.

\begin{table}[h!]
\centering
\begin{tabular}{|c||cc|}
\hline
Model                       & \# params (millions) & Accuracy (\%) \\ \hline \hline
\multicolumn{2}{|c|}{ResNet \cite{DBLP:journals/corr/HeZRS15}}                       & our/\textit{original}  \\ \hline
ResNet-110                   & 1.73           & 94.09/\textit{93.57}   \\
ResNet-56                    & 0.86          & 93.61/\textit{93.03}   \\
ResNet-44                    & 0.66                 		& 93.21/\textit{92.83}   \\
ResNet-32                    & 0.47           & 92.91/\textit{92.49}   \\
ResNet-20                    & 0.27                 		& 92.19/\textit{91.25}   \\ \hline \hline
\multicolumn{3}{|c|}{Budgeted ResNet}                              \\ \hline
\multirow{7}{*}{B-ResNet}   & 4.38                 & 94.35         \\
                            & 2.27                 & 94.2          \\
                            & 1.29                 & 93.85         \\
                            & 0.48                 & 93.42         \\
                            & 0.34                 & 92.72         \\
                            & 0.3                  & 92.52         \\
                            & 0.29                 & 92.17         \\ \hline \hline
\multicolumn{2}{|c|}{Convolutional Neural Fabrics \cite{DBLP:journals/corr/SaxenaV16}} & our/\textit{original}  \\ \hline
CNF W=8                     & 18.04           & 94.83/\textit{90.58}   \\
CNF W=4                     & 8.58              & 93.75/\textit{87.91}   \\
CNF W=2                     & 3.85              & 92.54/\textit{86.21}   \\
CNF W=1                     & 0.74                 & 89.91         \\ \hline \hline
\multicolumn{3}{|c|}{Budgeted CNF}                                 \\ \hline
\multirow{6}{*}{B-CNF}      & 7.56                 & 94.88         \\
                            & 4.98                 & 94.58         \\
                            & 4.28                 & 94.55         \\
                            & 3.67                 & 94.42         \\
                            & 2.65                 & 94.00         \\
                            & 1.19                 & 93.53         \\ \hline
\end{tabular}
\caption{Accuracy/memory trade-off on Cifar-10 using B-ResNet and B-CNF.}
\label{cif10_allmodels_params}
\end{table}
\vspace{-0.3cm}
\paragraph{Reducing the memory consumption:} Similar experiments have been made considering the \textit{memory consumption cost} that measures the number of parameters of the learned architectures. We want to demonstrate here the ability of our technique to be used with a large variety of costs, and not only to reduce the computation time. Table \ref{cif10_allmodels_params} illustrates the results obtained on CIFAR-10. As with the \textit{computation cost}, one can see that our approach is able to discover architectures that, given a particular memory cost, obtain a better accuracy. For example, for a model which size is $\approx 0.47$ millions parameters, ResNet-32 has a classification error of 7.81\%  while B-ResNet only makes 6.58\% error with $\approx 0.48$ million parameters.

\paragraph{Image Segmentation: }



We also perform experiments on the image segmentation task using the Part Label dataset with CNF and B-CNF (Table \ref{tab:partlabels}). In this task, the model computes a map of pixel probabilities, the output layer being now located at the top-right position of the CNF matrix. It is thus more difficult to reduce the overall computation cost. On the Part Label dataset, we are able to learn a BSN model with a computation gain of $40\%$. Forcing the model to reduce further the computation cost by decreasing the value of $\mathbf{C}$ results in inconsistent models. At a computation gain of $40 \%$, BSN obtains an error rate of $4.57 \%$, which can be compared with the error of $4.94\%$ for the full model. The B-CNF best learned architecture is given in appendix.
\vspace{-0.2cm}
\paragraph{Learning Dynamics: }

\begin{figure}[ht]
\begin{center}
\includegraphics[width=0.8\linewidth]{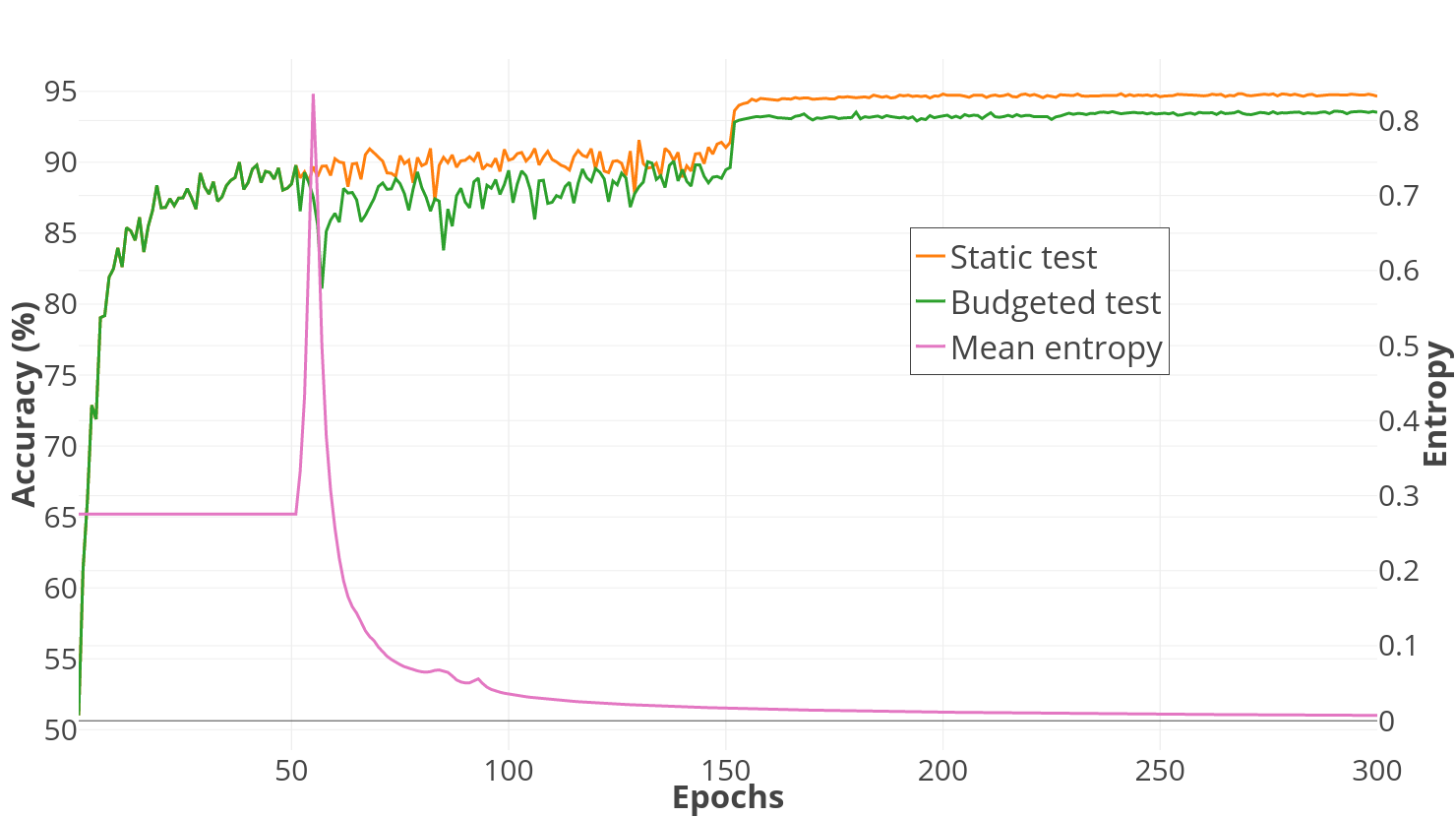}
\end{center}

\caption{Evolution of the loss function and the entropy of $\Gamma$ during training. The period between epoch 0 and 50 is the burn-in phase. The learning rate is divided by 10 after epoch 150 to increase the convergence speed.}
\label{fig:dynamics}
\end{figure}

Figure \ref{fig:dynamics} illustrates the learning dynamics of B-CNF and CNF. First, one can see (entropy curve) that the model becomes deterministic at the end of the learning procedure, and thus converges to a unique architecture. Moreover, the training speed of B-CNF and CNF are comparable showing that our method does not result in a slower training procedure. Note that the figure illustrates the fact that during a burn-in period, we don't update the edges probabilities, which allows us to obtain a faster convergence speed (see appendix).

\begin{table}[h]
\small{
\centering
\begin{tabular}{|c||cc|}
\hline
Model                       & FLOPs(billions) & Accuracy \\ \hline \hline
CNF                         & 35.614          & 95.06    \\ \hline
CNF W=8 \cite{DBLP:journals/corr/SaxenaV16} & 35.614          & 95.39    \\ \hline
B-CNF                       & 28.49           & 95.21    \\
B-CNF                       & 21.37           & 95.43    \\ \hline
\end{tabular}
}
\caption{Accuracy/Speed trade-off on Part Label using CNF.}
\label{tab:partlabels}
\end{table}

\subsection{Learning Distributed Architectures}\label{parallel_section}

\begin{figure*}[h]
\centering
    \begin{subfigure}[t]{0.65\textwidth}
        \includegraphics[width=1.0\linewidth]{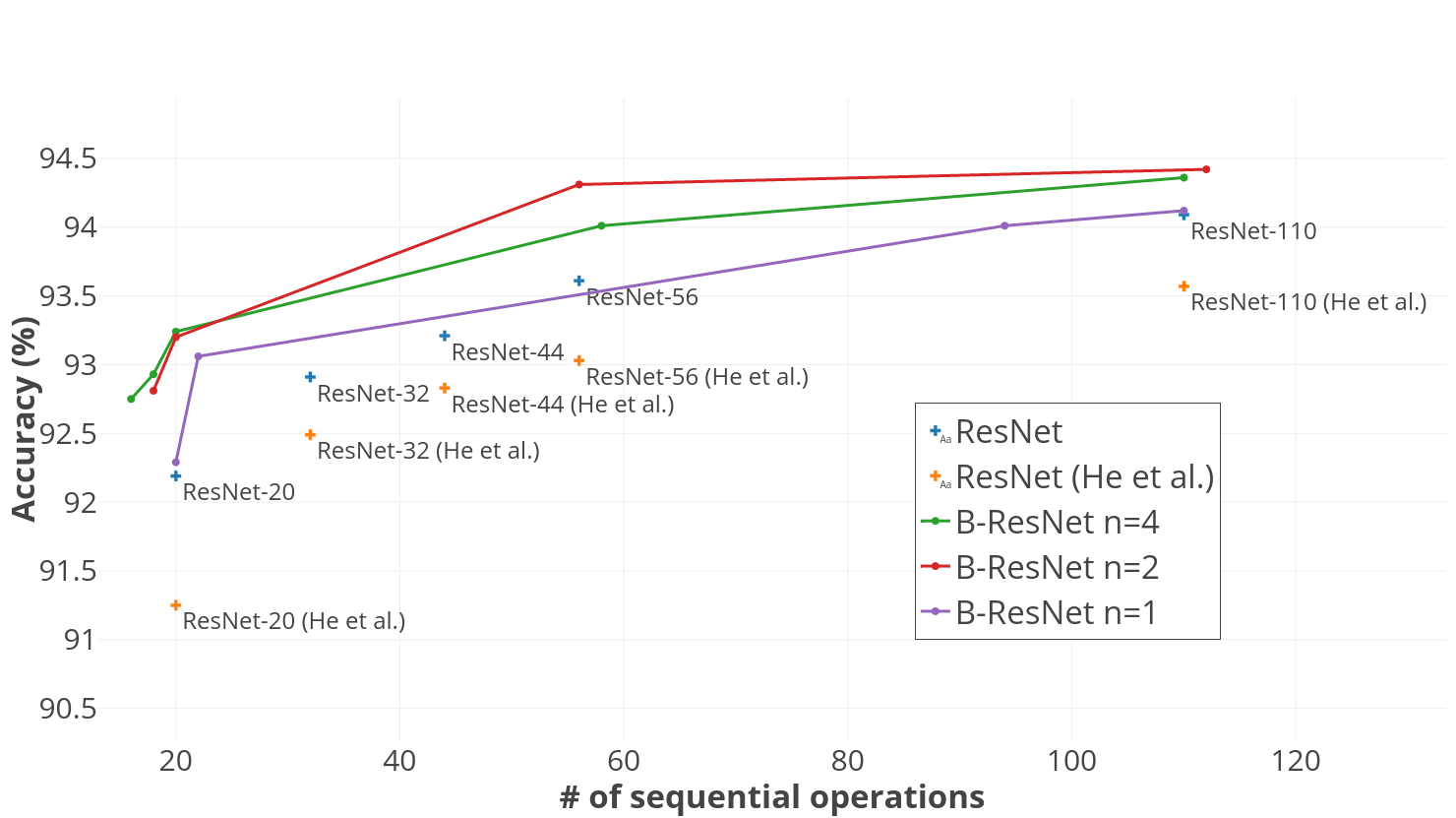}
        \caption{Accuracy/number of operation for different number of cores on CIFAR-10 using B-ResNet.}
     \end{subfigure} 
     \hspace{.7em}
     \begin{subfigure}[t]{0.30\textwidth}
     	\vspace{-15em}
        \includegraphics[width=1.0\linewidth]{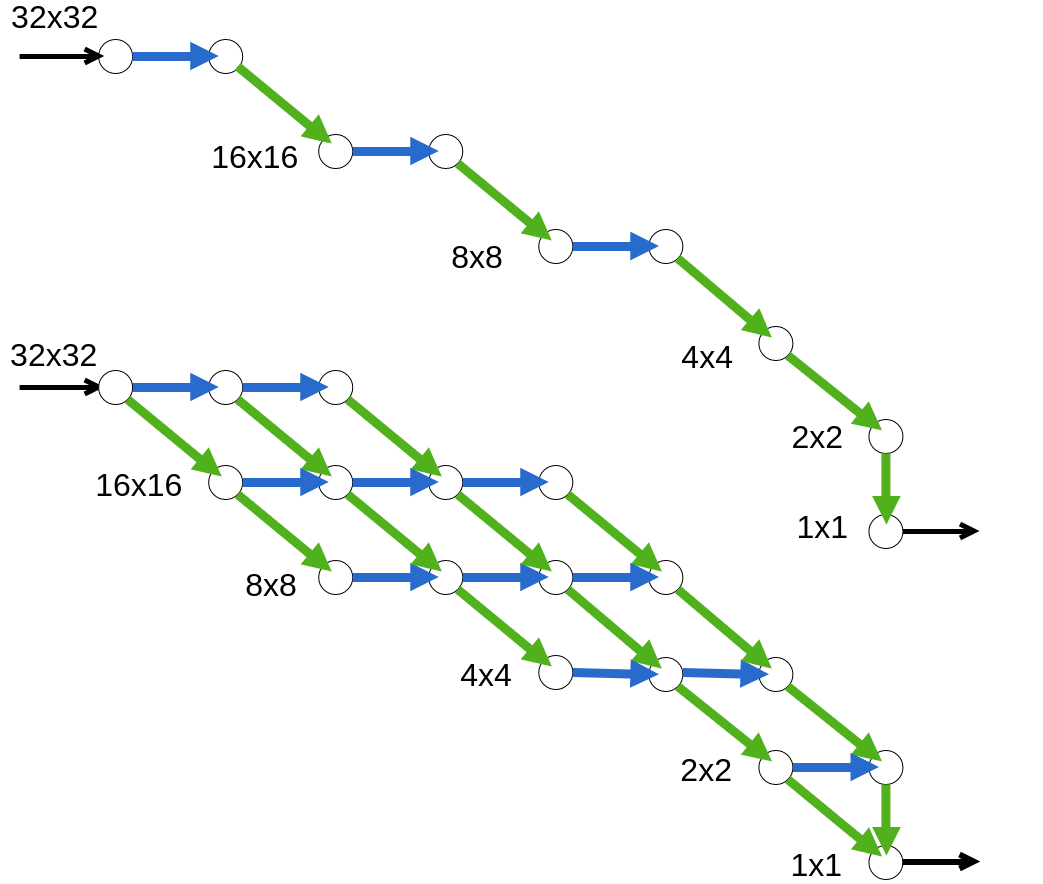}
        \caption{Architectures discovered with B-CNF for different number of cores: $n=1$ (top) and $n=4$ (bottom)}
        \label{fig:para_arch}
     \end{subfigure}
\caption{Architectures discovered on CIFAR-10 for different number of distributed cores $n$.}
\label{fig:para_res_arch}
\end{figure*}



At last, we perform a third set of experiments focused on distributed computing where different edges can be computed simultaneously on different computers/processors of a distributed platform. We thus evaluate the quality of an architecture by its ability to be efficiently parallelized. The \textit{distributed computation cost} corresponds to the number of steps needed to compute the output of the network e.g on an architecture with $n=2$ computers, depending on the structure of the network, two edges could be computed simultaneously. If the architecture is a sequence of layers, then this parallelization becomes impossible. Theses experiments allow us to measure the ability of BSN to handle complex costs that cannot be decomposed as a sum of individual modules costs as it is usually done in related works. 

Results and corresponding architectures are illustrated in Figure \ref{fig:para_res_arch} for the CIFAR-10 dataset and for both the B-ResNet and the B-CNF architectures. Note that ResNet is typically an architecture that cannot be efficiently distributed since it is a sequences of modules. One can see that our approach is able to find efficient architectures for $n=2$ and $n=4$. Surprisingly, when $n=4$ the discovered architectures are less efficient, which is mainly due to an over fitting of the training set, the cost constraint becomes too large and stop acting as a regularizer on the network architecture. On Figure \ref{fig:para_arch}, one can see two examples of architectures discovered when $n=1$ and $n=4$. The shape of the architecture when $n=4$ clearly confirm that BSN is able to discover parallelized architectures, and to 'understand' the structure of this complex cost.  

\section{Related Work}\label{sec_related_work}

\textbf{Learning cost-efficient models: } One of the first approaches to learn efficient models is to \textit{a posteriori}  compress the learned network, typically by pruning some connections. The oldest work is certainly the Optimal Brain Surgeon \cite{Hassibi} which removes weights in a classical neural network. The problem of network compression can also be seen as a way to speed up a particular architecture, for example by using quantization of the weights of the network \cite{Vanhoucke11}, or by combining pruning and quantization  \cite{han}. Other algorithms include the use of hardware efficient operations that allow a high speedup \cite{hard}. 

 \textbf{Efficient architectures: }Architecture improvements have been widely used in CNN to improve cost efficiency of network components, some examples are the bottleneck units in the ResNet model \cite{DBLP:journals/corr/HeZRS15}, the use of depthwise separable convolution in Xception \cite{DBLP:journals/corr/Chollet16a} and the lightweight  MobileNets\cite{DBLP:journals/corr/HowardZCKWWAA17} or the combinaison of pointwise group convolution and channel shuffle in ShuffleNet\cite{DBLP:journals/corr/ZhangZLS17}. 

\textbf{End-to-end approaches: } A first example of end-to-end approaches is the usage of quantization at training time: different authors trained models using binary weight quantization coupled with full precision arithmetic operations  \cite{DBLP:journals/corr/CourbariauxBD15},\cite{DBLP:journals/corr/Lu17c}. Recently, \cite{DBLP:journals/corr/abs-1710-03740} proposed an method using half precision floating numbers during training. Another technique proposed by \cite{DBLP:journals/corr/HintonVD15}, \cite{DBLP:journals/corr/RomeroBKCGB14} and used in \cite{2017arXiv170510924Z,2017arXiv170510194N} is the distillation of knowledge, which consists of training a smaller network to imitate the outputs of a larger network. 
Other approaches are dynamic networks which conditionally select the modules to respect a budget objective.\cite{DBLP:journals/corr/BolukbasiWDS17, DBLP:journals/corr/OdenaLO17,DBLP:journals/corr/HuangCLWMW17,DBLP:journals/corr/BengioBPP15,DBLP:journals/corr/McGillP17}. 

\textbf{Architecture Search: } Different authors have proposed to provide networks with the ability to learn to select the computations that will be applied i.e choosing the right architecture for a particular task. This is the case for example for classification in \cite{DBLP:journals/corr/DenoyerG14,DBLP:journals/corr/ZophL16} based on Reinforcement learning techniques, in \cite{DBLP:journals/corr/SrivastavaGS15} based on gating mechanisms,  in \cite{DBLP:journals/corr/RealMSSSLK17} based on  evolutionary algorithms or even in \cite{DBLP:journals/corr/FernandoBBZHRPW17} based on both RL and evolutionary techniques.

The strongest difference w.r.t. existing methods is that we do not make any assumption concerning the nature of the cost. Our model is thus more generic than existing techniques and allow to handle a large variety of problems.

\section{Conclusion and Perspectives}

We proposed a new model called \textit{Budgeted Super Network} able to automatically discover cost-constrained neural network architectures by specifying a maximum authorized cost. The experiments in the computer vision domain show the effectiveness of our approach. Its main advantage is that BSN can be used for any costs (computation cost, memory cost, etc.) without any assumption on the shape of this cost. A promising research direction is now to study whether this model could be adapted in order to reduce the training time (instead of the test computation time). This could for example be done using meta-learning approaches.

\section*{Acknowledgments}
This work has been funded in part by grant ANR-16-CE23-0016 ``PAMELA'' and grant ANR-16-CE23-0006 ``Deep in France''.

{\small
\bibliographystyle{ieee}
\bibliography{egbib}

\begin{thebibliography}{10}\itemsep=-1pt

\bibitem{DBLP:journals/corr/BengioBPP15}
E.~Bengio, P.~Bacon, J.~Pineau, and D.~Precup.
\newblock Conditional computation in neural networks for faster models.
\newblock {\em CoRR}, abs/1511.06297, 2015.

\bibitem{DBLP:journals/corr/BolukbasiWDS17}
T.~Bolukbasi, J.~Wang, O.~Dekel, and V.~Saligrama.
\newblock Adaptive neural networks for fast test-time prediction.
\newblock {\em CoRR}, abs/1702.07811, 2017.

\bibitem{DBLP:journals/corr/Chollet16a}
F.~Chollet.
\newblock Xception: Deep learning with depthwise separable convolutions.
\newblock {\em CoRR}, abs/1610.02357, 2016.

\bibitem{DBLP:journals/corr/CourbariauxBD15}
M.~Courbariaux, Y.~Bengio, and J.~David.
\newblock Binaryconnect: Training deep neural networks with binary weights
  during propagations.
\newblock {\em CoRR}, abs/1511.00363, 2015.

\bibitem{DBLP:journals/corr/DenoyerG14}
L.~Denoyer and P.~Gallinari.
\newblock Deep sequential neural network.
\newblock {\em CoRR}, abs/1410.0510, 2014.

\bibitem{hard}
J.~Devlin.
\newblock {Sharp Models on Dull Hardware: Fast and Accurate Neural Machine
  Translation Decoding on the CPU}, 2017.

\bibitem{DBLP:journals/corr/DongHYY17}
X.~Dong, J.~Huang, Y.~Yang, and S.~Yan.
\newblock More is less: {A} more complicated network with less inference
  complexity.
\newblock {\em CoRR}, abs/1703.08651, 2017.

\bibitem{DBLP:journals/corr/FernandoBBZHRPW17}
C.~Fernando, D.~Banarse, C.~Blundell, Y.~Zwols, D.~Ha, A.~A. Rusu, A.~Pritzel,
  and D.~Wierstra.
\newblock Pathnet: Evolution channels gradient descent in super neural
  networks.
\newblock {\em CoRR}, abs/1701.08734, 2017.

\bibitem{han}
S.~Han, H.~Mao, and W.~J. Dally.
\newblock Deep compression: Compressing deep neural networks with pruning,
  trained quantization and huffman coding.
\newblock In {\em International Conference on Learning Representations (ICLR'16
  best paper award)}, 2015.

\bibitem{Hassibi}
B.~Hassibi and D.~G. Stork.
\newblock Second order derivatives for network pruning: Optimal brain surgeon.
\newblock In {\em Advances in Neural Information Processing Systems 5, [NIPS
  Conference]}, pages 164--171, San Francisco, CA, USA, 1993. Morgan Kaufmann
  Publishers Inc.

\bibitem{DBLP:journals/corr/HeZRS15}
K.~He, X.~Zhang, S.~Ren, and J.~Sun.
\newblock Deep residual learning for image recognition.
\newblock {\em CoRR}, abs/1512.03385, 2015.

\bibitem{DBLP:journals/corr/HintonVD15}
G.~E. Hinton, O.~Vinyals, and J.~Dean.
\newblock Distilling the knowledge in a neural network.
\newblock {\em CoRR}, abs/1503.02531, 2015.

\bibitem{DBLP:journals/corr/HowardZCKWWAA17}
A.~G. Howard, M.~Zhu, B.~Chen, D.~Kalenichenko, W.~Wang, T.~Weyand,
  M.~Andreetto, and H.~Adam.
\newblock Mobilenets: Efficient convolutional neural networks for mobile vision
  applications.
\newblock {\em CoRR}, abs/1704.04861, 2017.

\bibitem{DBLP:journals/corr/HuangCLWMW17}
G.~Huang, D.~Chen, T.~Li, F.~Wu, L.~van~der Maaten, and K.~Q. Weinberger.
\newblock Multi-scale dense convolutional networks for efficient prediction.
\newblock {\em CoRR}, abs/1703.09844, 2017.

\bibitem{GLOC_CVPR13}
A.~Kae, K.~Sohn, H.~Lee, and E.~Learned-Miller.
\newblock Augmenting {CRF}s with {B}oltzmann machine shape priors for image
  labeling.
\newblock In {\em CVPR}, 2013.

\bibitem{Krizhevsky09learningmultiple}
A.~Krizhevsky.
\newblock Learning multiple layers of features from tiny images.
\newblock Technical report, 2009.

\bibitem{DBLP:journals/corr/Lu17c}
L.~Lu.
\newblock Toward computation and memory efficient neural network acoustic
  models with binary weights and activations.
\newblock {\em CoRR}, abs/1706.09453, 2017.

\bibitem{DBLP:journals/corr/McGillP17}
M.~McGill and P.~Perona.
\newblock Deciding how to decide: Dynamic routing in artificial neural
  networks.
\newblock {\em CoRR}, abs/1703.06217, 2017.

\bibitem{DBLP:journals/corr/abs-1710-03740}
P.~Micikevicius, S.~Narang, J.~Alben, G.~F. Diamos, E.~Elsen, D.~Garcia,
  B.~Ginsburg, M.~Houston, O.~Kuchaiev, G.~Venkatesh, and H.~Wu.
\newblock Mixed precision training.
\newblock {\em CoRR}, abs/1710.03740, 2017.

\bibitem{DBLP:journals/corr/MiikkulainenLMR17}
R.~Miikkulainen, J.~Z. Liang, E.~Meyerson, A.~Rawal, D.~Fink, O.~Francon,
  B.~Raju, H.~Shahrzad, A.~Navruzyan, N.~Duffy, and B.~Hodjat.
\newblock Evolving deep neural networks.
\newblock {\em CoRR}, abs/1703.00548, 2017.

\bibitem{2017arXiv170510194N}
F.~{Nan} and V.~{Saligrama}.
\newblock {Adaptive Classification for Prediction Under a Budget}.
\newblock {\em ArXiv e-prints}, May 2017.

\bibitem{DBLP:journals/corr/OdenaLO17}
A.~Odena, D.~Lawson, and C.~Olah.
\newblock Changing model behavior at test-time using reinforcement learning.
\newblock {\em CoRR}, abs/1702.07780, 2017.

\bibitem{DBLP:journals/corr/RealMSSSLK17}
E.~Real, S.~Moore, A.~Selle, S.~Saxena, Y.~L. Suematsu, Q.~V. Le, and
  A.~Kurakin.
\newblock Large-scale evolution of image classifiers.
\newblock {\em CoRR}, abs/1703.01041, 2017.

\bibitem{DBLP:journals/corr/RomeroBKCGB14}
A.~Romero, N.~Ballas, S.~E. Kahou, A.~Chassang, C.~Gatta, and Y.~Bengio.
\newblock Fitnets: Hints for thin deep nets.
\newblock {\em CoRR}, abs/1412.6550, 2014.

\bibitem{DBLP:journals/corr/SaxenaV16}
S.~Saxena and J.~Verbeek.
\newblock Convolutional neural fabrics.
\newblock {\em CoRR}, abs/1606.02492, 2016.

\bibitem{DBLP:journals/corr/SrivastavaGS15}
R.~K. Srivastava, K.~Greff, and J.~Schmidhuber.
\newblock Highway networks.
\newblock {\em CoRR}, abs/1505.00387, 2015.

\bibitem{DBLP:conf/gecco/StanleyM02a}
K.~O. Stanley and R.~Miikkulainen.
\newblock Efficient reinforcement learning through evolving neural network
  topologies.
\newblock In {\em {GECCO} 2002: Proceedings of the Genetic and Evolutionary
  Computation Conference, New York, USA, 9-13 July 2002}, pages 569--577, 2002.

\bibitem{DBLP:journals/corr/SzegedyLJSRAEVR14}
C.~Szegedy, W.~Liu, Y.~Jia, P.~Sermanet, S.~E. Reed, D.~Anguelov, D.~Erhan,
  V.~Vanhoucke, and A.~Rabinovich.
\newblock Going deeper with convolutions.
\newblock {\em CoRR}, abs/1409.4842, 2014.

\bibitem{Vanhoucke11}
V.~Vanhoucke, A.~Senior, and M.~Z. Mao.
\newblock Improving the speed of neural networks on cpus.
\newblock In {\em Deep Learning and Unsupervised Feature Learning Workshop,
  NIPS 2011}, 2011.

\bibitem{DBLP:journals/corr/ZhangZLS17}
X.~Zhang, X.~Zhou, M.~Lin, and J.~Sun.
\newblock Shufflenet: An extremely efficient convolutional neural network for
  mobile devices.
\newblock {\em CoRR}, abs/1707.01083, 2017.

\bibitem{2017arXiv170510924Z}
H.~{Zhu}, F.~{Nan}, I.~{Paschalidis}, and V.~{Saligrama}.
\newblock {Sequential Dynamic Decision Making with Deep Neural Nets on a
  Test-Time Budget}.
\newblock {\em ArXiv e-prints}, May 2017.

\bibitem{DBLP:journals/corr/ZophL16}
B.~Zoph and Q.~V. Le.
\newblock Neural architecture search with reinforcement learning.
\newblock {\em CoRR}, abs/1611.01578, 2016.

\end{thebibliography}
}

\newpage
\clearpage

\thispagestyle{empty}
\section*{Supplementary Material}

\subsection*{Demonstration of Proposition 1}

Let us consider the stochastic optimization problem defined in Equation \ref{stochobjective}. The schema of the proof is the following:
\begin{itemize}
\item First, we lower bound the value of Equation \ref{stochobjective} by the optimal value of Equation \ref{objective}.
\item Then we show that this lower bound can be reached by some particular values of $\Gamma$ and $\theta$ in Equation \ref{stochobjective}. Said otherwise, the solution of Equation \ref{stochobjective} is equivalent to the solution of \ref{objective}.
\end{itemize}

Let us denote:
\begin{multline}
B(H \odot E, \theta,\lambda) = \frac{1}{\ell} \sum_i \Delta(f(x^i,H \odot E, \theta),y^i)  \\+ \lambda \max (0,C(H \odot E)-C)
\end{multline}

Given a value of $\Gamma$, let us denote $supp(\Gamma)$ all the $H$ matrices that can be sampled following $\Gamma$. The objective function of Equation \ref{stochobjective} can be written as:
\begin{equation}
\begin{aligned}
E_{H \sim \Gamma} [B(H \odot E, \theta,\lambda)] &=\sum\limits_{H \in supp(\Gamma)} B(H \odot E, \theta,\lambda) P(H|\Gamma) \\
& \geq \sum\limits_{H \in supp(\Gamma)} B((H \odot E)^*, \theta^*,\lambda) P(H|\Gamma) \\
& = B((H \odot E)^*, \theta^*,\lambda)
\end{aligned}
\label{cucu}
\end{equation}
where $(H \odot E)^*$ and $\theta^*$ correspond to the solution of:
\begin{equation}
(H \odot E)^*,\theta^* = \arg \min_{H,\theta} B(H \odot E,\theta,\lambda)
\label{eqt}
\end{equation}

Now, it is easy to show that this lower bound can be reached by considering a value of $\Gamma^*$ such that $\forall H \in supp(\Gamma), H \odot E = (H \odot E)^*$. This corresponds to a value of $\Gamma$ where all the probabilities associated to edges in $E$ are equal to $0$ or to $1$.

\newpage

\subsection*{Gradient computation}



\begin{equation}
\nabla_{\theta,\Gamma} \mathcal{L}(x,y,E,\Gamma,\theta) = \nabla_{\theta,\Gamma} \mathbb{E}_{H \sim \Gamma} \mathcal{D}(x,y,\theta,E,H)
\end{equation}

\begin{equation}
= \nabla_{\theta,\Gamma} \sum\limits_H P(H|\Gamma) \mathcal{D}(x,y,\theta,E,H)
\end{equation}

\begin{equation}
= \sum\limits_H \nabla_{\theta,\Gamma} (P(H|\Gamma) \mathcal{D}(x,y,\theta,E,H))
\end{equation}

\begin{multline}
= \sum\limits_H \nabla_{\theta,\Gamma} (P(H|\Gamma))\mathcal{D}(x,y,\theta,E,H) \\+ P(H|\Gamma) \nabla_{\theta,\Gamma} \mathcal{D}(x,y,\theta,E,H)
\end{multline}

\begin{multline}
= \sum\limits_H P(H|\Gamma)\nabla_{\theta,\Gamma} \log P(H|\Gamma) \mathcal{D}(x,y,\theta,E,H) \\+ P(H|\Gamma) \nabla_{\theta,\Gamma} \mathcal{D}(x,y,\theta,E,H)
\end{multline}

Using Equation \ref{D_def}:

\begin{multline}
= \sum\limits_H P(H|\Gamma) ((\nabla_{\theta,\Gamma} \log P(H|\Gamma)) \mathcal{D}(x,y,\theta,E,H) \\+ \nabla_{\theta,\Gamma} \Delta(f(x,H \odot E,\theta),y) )
\end{multline}

\begin{multline}
= \sum\limits_H P(H|\Gamma) \left[(\nabla_{\theta,\Gamma}  \log P(H|\Gamma)) \mathcal{D}(x,y,\theta,E,H) \right] \\+ \sum\limits_H P(H|\Gamma) \left[ \nabla_{\theta,\Gamma} \Delta(f(x,H \odot E,\theta),y) \right]
\end{multline}

\begin{multline}
= \sum\limits_H P(H|\Gamma) \left[ (\nabla_{\theta,\Gamma}  \log P(H|\Gamma)) \Delta(f(x,H \odot E,\theta),y) \right]\\
+ \lambda \sum\limits_H P(H|\Gamma) \left[ (\nabla_{\theta,\Gamma}  \log P(H|\Gamma)) \max(0, C(H \odot E) - \mathbf{C}) \right] \\
+ \sum\limits_H P(H|\Gamma) \left[ \nabla_{\theta,\Gamma} \Delta(f(x,H \odot E,\theta),y) \right]
\end{multline}

\newpage

\subsection*{Segmentation architecture}
\begin{figure}[ht]
\centering
\includegraphics[width=\linewidth]{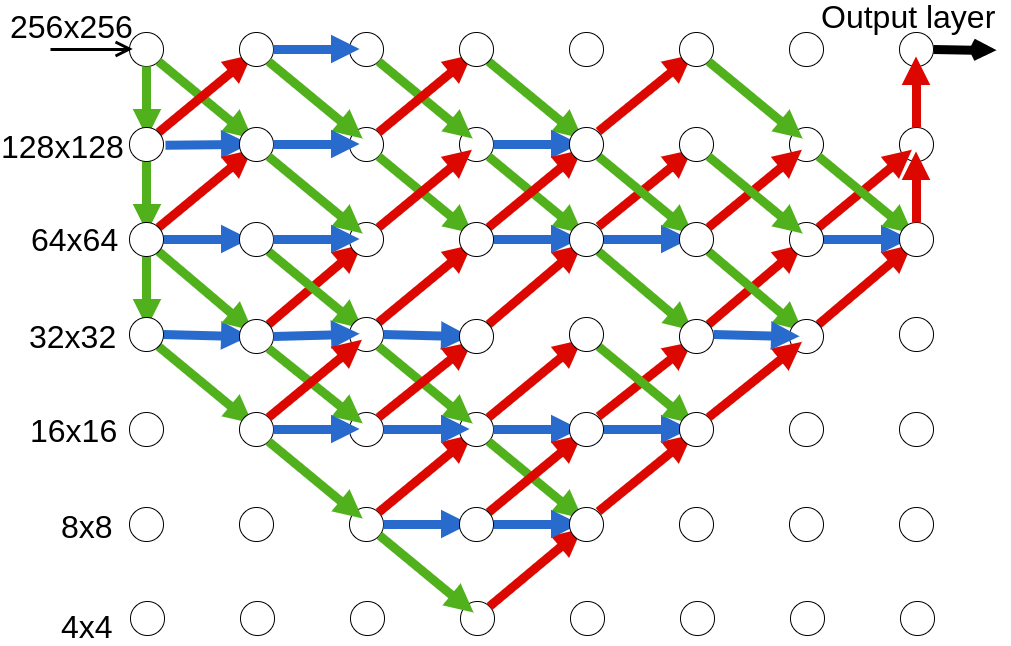}
\caption{Segmentation architecture}
\label{fig:partlabel_arch}
\end{figure}
Figure \ref{fig:partlabel_arch} is an example of segmentation architecture discovered on the Part Label dataset using the \textit{flop cost}. It is interesting to note that only one layer with 256x256 input and output is kept and that most of the computations are done at lower less-expensive layers.

\subsection*{Model Selection Protocol}
\begin{figure}[ht]
\centering
\includegraphics[width=1.1\linewidth]{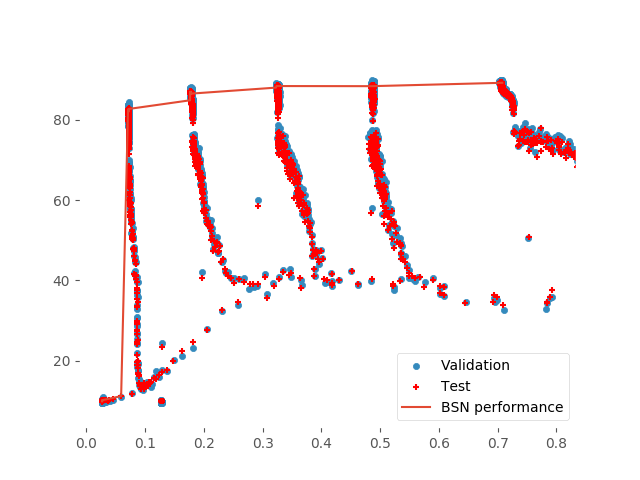}
\caption{Model selection}
\label{fig:model_selection}
\end{figure}

The selection of reported models is obtained by learning many different models, computing the Pareto front of the accuracy/cost curve on the validation set, and reporting the performance obtained on the test set. This is illustrated in figure \ref{fig:model_selection} where many different models are reported on the validation set(blue circles) with the corresponding performance on the test set (red crosses).

\newpage

\section*{Considering non-differentiable costs}
\subsection*{Stochastic costs in the REINFORCE algorithm: } As explained previously, the proposed algorithm can also be used when the cost $C(H \odot E)$ is a stochastic function that depends on the environment e.g the network latency, (or even on the input data $x$). Our algorithm is still able to learn with such stochastic costs since the only change in the learning objective is that the expectation is now made on both $H$ and $C$ (and $x$ if needed). 
This property is interesting since it allows to discover efficient architecture on stochastic operational infrastructure. 


\paragraph*{Distributed computation cost} Taking the real-life example of a network which will, once optimized, have to run on a given computing infrastructure, the \textit{distributed computation cost} is a measure of how "parallelizable" an architecture is. This cost function takes the following three elements as inputs (i)A network architecture (represented as a graph for instance), (ii)An allocation algorithm and (iii) a maximum number of concurrent possible operations. The cost function then returns the number of computation cycles required to run the architecture given the allocation strategy.
\begin{figure}[ht]
	\centering
    \includegraphics[width=9cm]{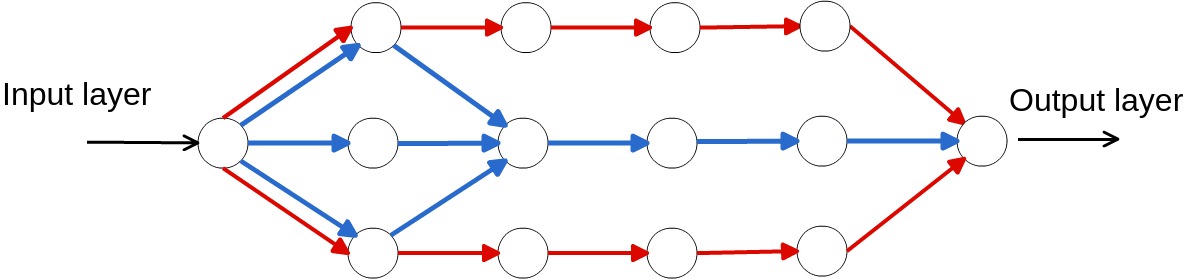}
	\caption{Two networks illustrating the need to have a cost function evaluating the global architecture of a network. Considering an environment with $n=2$ machines performing computations in parallel, the blue network composed of 9 computational modules has a \textit{distributed computation cost} of 6 while the red network, composed of 10 modules, has a smaller cost of 5.}
\label{parallel_ex}
\end{figure}


\subsection*{Additional Architecture Details}
\subsubsection*{ResNet Fabric}
Based on the ResNet architecture, the structure of a ResNet Fabric is a stack of $k$ groups of layers, each group being composed of $2n$ layers where $n$ represents the width of the Fabric. The feature maps size and number of filters stay constant across the layers of each group and are modified between groups.

Due to its linear structure, the standard ResNet architecture spans a limited number of possible (sub-)architectures. In order to increase the size of the search space, we add several connections between groups as shown in \ref{fig1a}: each block in the second to last groups receives two (for the first and last block of each group) or three (for every other block) inputs from preceding groups. To stay consistent with the rest of the network, each connection is a \textit{basic block} \cite{DBLP:journals/corr/HeZRS15} composed of 2 convolutional layers and a shortcut connection.

In our experiments, we use stacks of $k=3$ blocks and $n=\{3, 5, 7, 9, 18\}$ to respectively include the ResNet-$\{20,32,44,56,110\}$ in the Fabric. Between each block, the feature maps size is reduced by a factor of 2 and the number of feature maps is doubled.

\subsubsection*{Convolutional Neural Fabric}
The second network we use in our experiments is based on the dense Convolutional Neural Fabrics, which can be seen as a multi-layer and multi-scale convolutional neural network.  As shown in Figure \ref{fig1b}, this architecture has 2 axis: The first axis represents the different columns (or width) $W$ of the network while the second axis corresponds to different scales (or height) $H$ of output feature maps, the first scale being the size of the input images, each subsequent scale being of a size reduced by a factor of 2 up to the last scale corresponding to a single scalar.

Each layer $(l, s)$ in this fabric takes its input from three different layers of the preceding column: (i) One with a finer scale $(l-1, s-1)$ on which a convolution with stride 2 is applied to obtain feature maps having half the size of the input, (ii) one with the same scale $(l-1, s)$ on which a convolution with stride 1 is applied to obtain feature map of the same resolution as the input and (iii) one with a coarser scale $(l-1, s+1)$ on which  convolution with stride 1 is applied after a factor 2 up-sampling to obtain feature maps having twice the size of the input. The three feature blocks are then added before passing through the ReLU activation function to obtain the final output of this layer $(l, s)$. 

The first and last columns are the only two which have vertical connections within scales of the same layer (as can be seen in Figure \ref{fig1b}). This is made to allow the propagation of the information to all nodes in the first column and to aggregate the activations of the last column to compute the final prediction. A more detailed description of this architecture can be found in the CNF original article.

We used two different Convolutional Neural Fabrics in our experiments: One for the classification task (CIFAR-10 and CIFAR-100) with $W=8$ columns, $H=6$ scales and 128 filters per convolution and one for the segmentation task (Part Label) with $W=8$ layers, $H=9$ scales (from 256x256 to 1x1 feature map sizes) and 64 filters per convolution.

\newpage
\subsection*{Additional Learning Details}

\subsubsection*{Datasets}

\paragraph*{CIFAR-10.}
The CIFAR-10 dataset consists of 60k 32x32 images with 10 classes and 6000 images per class. The dataset is decomposed in 50k training and 10k testing images. We split the training set following the standard, i.e 45k training samples and 5k validation samples. We use two data augmentation techniques: padding the image to 36x36 pixels before extracting a random crop of size 32x32 and horizontally flipping. Images are then normalized in the range [-1,1].  

\paragraph*{CIFAR100.}
The CIFAR-100 dataset is similar to CIFAR-10, with 100 classes and 600 images per class. We use the same train/validation split and data augmentation technique as with CIFAR-10.  

\paragraph*{Part Labels.}
The Part Labels dataset is a subset of the LFW dataset composed of 2927 250x250 face images in which each pixel is labeled as one of the Hair/Skin/Background classes. The standard split contains 1500 training samples, 500 validation samples and 927 test samples. Images are zero-padded from 250x250 to 256x256. We use horizontal flipping as data augmentation. Images are then normalized in the range [-1,1].

\subsubsection*{Learning procedure}
When training our budgeted models, we first train the network for 50 "warm-up" epochs during which no sampling is done (The whole super network is trained). After this warm-up phase, the probability of each edge is initialized and we start sampling architectures.

The real-valued distribution parameter associated with each layer (and used to generate the probability of sampling the edge) are all initialized to 3, resulting in a $\approx 0.95$ initial probability once passed through the sigmoid activation function.

On CIFAR-10 and CIFAR-100 datasets we train all models for 300 epochs. We start with a learning rate of $10^{-1}$ and divide it by 10 after 150 and 225 epochs. On Part Label dataset all models are trained for 200 epochs with a learning rate initialized to $10^{-1}$ and divided by 10 after 130 epochs.

For all models and all cost functions, we select the $\lambda$ hyper-parameter based on the order of magnitude $m$ of the maximum authorized cost $\mathbf{C}$. $\lambda$ is determined using cross-validation on values logarithmically spaced between $10^{m-1}$ and $10^{m+1}$.

\newpage

\subsection*{Forward algorithm}
Given the SS-Network $(E,\Gamma, \theta)$ and input $x$, the evaluation of $f(x,E, \Gamma,\theta)$ is done as follow :

\begin{algorithm}[ht] 
\caption{Stochastic Super Network forward algorithm}
\begin{algorithmic}[1]
\Procedure{SSN-forward}{$x,E,\Gamma,\theta$}
\State $H \sim \Gamma$ \Comment{as explained in Section \ref{secion_ssn}}
\For{$i \in [1..N]$}
\State $l_i \gets $\O 
\EndFor
\State $l_1 \gets x$
\For{$i \in [2..N]$}
\State $l_i \gets \sum\limits_{k<i} e_{k,i} h_{k,i} f_{k,i}(l_k)$
\EndFor 
\State \textbf{return} $l_N$
\EndProcedure
\end{algorithmic}
\label{algo2}
\end{algorithm}

\subsection*{Additional results}

\begin{table}[h]
\centering
\begin{tabular}{|c||cc|}
\hline
Model                       & \# of sequential operations & Accuracy \%  \\ \hline \hline
\multicolumn{2}{|c|}{ResNet \cite{DBLP:journals/corr/HeZRS15}}                              & our/\textit{original} \\ \hline
ResNet-110                  & 110.00                      & 94.09/\textit{93.57}  \\
ResNet-56                   & 56.00                       & 93.61/\textit{93.03}  \\
ResNet-44                   & 44.00                       & 93.21/\textit{92.83}  \\
ResNet-32                   & 32.00                       & 92.91/\textit{92.49}  \\
ResNet-20                   & 20.00                       & 92.19/\textit{91.25}  \\ \hline \hline
\multicolumn{3}{|c|}{Busgeted ResNet}                                    \\ \hline
\multirow{5}{*}{B-ResNet}                   & 110.00                      & 94.36        \\
		                    & 58.00                       & 94.01        \\
		                    & 20.00                       & 93.24        \\
		                    & 18.00                       & 92.93        \\
		                    & 16.00                       & 92.75        \\ \hline \hline
\multicolumn{2}{|c|}{Convolutional Neural Fabric \cite{DBLP:journals/corr/SaxenaV16}}         & our/\textit{original} \\ \hline
CNF W=8                     & 53.00                       & 94.83/\textit{90.58}  \\
CNF W=4                     & 31.00                       & 93.75/\textit{87.91}  \\
CNF W=2                     & 19.00                       & 92.54/\textit{86.21}  \\
CNF W=1                     & 12.00                       & 89.91        \\ \hline \hline
\multicolumn{3}{|c|}{Budgeted CNF}                                       \\ \hline
\multirow{4}{*}{B-ResNet}	& 31.00                       & 94.96        \\
		                    & 25.00                       & 94.72        \\
	                        & 21.00                       & 94.36        \\
	                        & 18.00                       & 93.86        \\ \hline
\end{tabular}
\caption{Results for \textit{Distributed computation cost} on CIFAR-10 with $n=4$}
\label{cif10_para4}
\end{table}

\begin{table}[h]
\centering
\begin{tabular}{|c||cc|}
\hline
Model                       & \# of sequential operations & Accuracy \%  \\ \hline \hline
\multicolumn{2}{|c|}{ResNet \cite{DBLP:journals/corr/HeZRS15}}                              & our/\textit{original} \\ \hline
ResNet-110                  & 112.00                      & 94.09/\textit{93.57}  \\
ResNet-56                   & 58.00                       & 93.61/\textit{93.03}  \\
ResNet-44                   & 46.00                       & 93.21/\textit{92.83}  \\
ResNet-32                   & 34.00                       & 92.91/\textit{92.49}  \\
ResNet-20                   & 22.00                       & 92.19/\textit{91.25}  \\ \hline \hline
\multicolumn{3}{|c|}{Budgeted ResNet}                                    \\ \hline
\multirow{5}{*}{B-ResNet}   & 184.00                      & 94.42        \\
                            & 110.00                      & 94.12        \\
                            & 94.00                       & 94.01        \\
                            & 22.00                       & 93.06        \\
                            & 20.00                       & 92.29        \\ \hline \hline
\multicolumn{2}{|c|}{Convolutional Neural Fabrics \cite{DBLP:journals/corr/SaxenaV16}}        & our/\textit{original} \\ \hline
CNF W=8                     & 171.00                      & 94.83/\textit{90.58}  \\
CNF W=4                     & 83.00                       & 93.75/\textit{87.91}  \\
CNF W=2                     & 39.00                       & 92.54/\textit{86.21}  \\
CNF W=1                     & 12.00                       & 89.91        \\ \hline \hline
\multicolumn{3}{|c|}{Budgeted CNF}                                       \\ \hline
\multirow{7}{*}{B-CNF}      & 98.00                       & 95.02        \\
                            & 50.00                       & 94.62        \\
                            & 45.00                       & 94.55        \\
                            & 39.00                       & 94.35        \\
                            & 33.00                       & 93.00        \\
                            & 26.00                       & 92.91        \\
                            & 18.00                       & 92.87        \\ \hline
\end{tabular}
\caption{Results for \textit{Distributed computation cost} on CIFAR-10 with $n=1$}
\label{cif10_para1}
\end{table}
\begin{table}[h]
\centering
\begin{tabular}{|c||cc|}
\hline
Model                         & \# of sequential operations & Accuracy (\%) \\ \hline \hline
\multicolumn{3}{|c|}{ResNet\cite{DBLP:journals/corr/HeZRS15}}               \\ \hline
ResNet-110                    & 112.00                      & 71.85         \\
ResNet-56                     & 58.00                       & 70.57         \\
ResNet-44                     & 46.00                       & 70.28         \\
ResNet-32                     & 34.00                       & 69.28         \\
ResNet-20                     & 22.00                       & 67.14         \\ \hline \hline
\multicolumn{3}{|c|}{Budgeted ResNet}             \\ \hline
\multirow{6}{*}{B-ResNet} & 320.00                      & 74.35         \\
                              & 184.00                      & 73.85         \\
                              & 110.00                      & 72.88         \\
                              & 67.00                       & 72.02         \\
                              & 32.00                       & 69.60         \\
                              & 20.00                       & 68.48         \\ \hline 
\end{tabular}
\caption{Results for \textit{Distributed computation cost} on CIFAR-100 with $n=1$}
\label{cif100_resnetfab_para1}
\end{table}

\begin{table}[h]
\centering
\begin{tabular}{|c||cc|}
\hline
Model                       & \# of sequential operations & Accuracy \%  \\ \hline \hline
\multicolumn{2}{|c|}{ResNet \cite{DBLP:journals/corr/HeZRS15}}                              & our/\textit{original} \\ \hline
ResNet-110                  & 110.00                      & 94.09/\textit{93.57}  \\
ResNet-56                   & 56.00                       & 93.61/\textit{93.03}  \\
ResNet-44                   & 44.00                       & 93.21/\textit{92.83}  \\
ResNet-32                   & 32.00                       & 92.91/\textit{92.49}  \\
ResNet-20                   & 20.00                       & 92.19/\textit{91.25}  \\ \hline \hline
\multicolumn{3}{|c|}{Budgeted ResNet}                                    \\ \hline
\multirow{5}{*}{B-ResNet}	& 179.00                      & 94.36        \\
        		            & 112.00                      & 94.42        \\
		                    & 56.00                       & 94.31        \\
				            & 20.00                       & 93.20        \\
				            & 18.00                       & 92.81        \\ \hline \hline
\multicolumn{2}{|c|}{Convolutional Neural Fabric \cite{DBLP:journals/corr/SaxenaV16}}         & our/\textit{original} \\ \hline
CNF W=8                     & 90.00                       & 94.83/\textit{90.58}  \\
CNF W=4                     & 47.00                       & 93.75/\textit{87.91}  \\
CNF W=2                     & 26.00                       & 92.54/\textit{86.21}  \\
CNF W=1                     & 12.00                       & 89.91        \\ \hline \hline
\multicolumn{3}{|c|}{Budgeted CNF}                                       \\ \hline
\multirow{6}{*}{B-CNF}      & 47.00                       & 94.67        \\
                            & 30.00                       & 94.68        \\
                            & 28.00                       & 94.58        \\
                            & 24.00                       & 94.41        \\
                            & 20.00                       & 94.35        \\
                            & 18.00                       & 92.86        \\ \hline
\end{tabular}
\caption{Results for \textit{Distributed computation cost} on CIFAR-10 with $n=2$}
\label{cif10_para2}
\end{table}
\begin{table}[]
\centering
\begin{tabular}{|c||cc|}
\hline
Model                     & \# of sequential operations & Accuracy (\%) \\ \hline \hline
\multicolumn{3}{|c|}{ResNe \cite{DBLP:journals/corr/HeZRS15}}                                            \\ \hline
ResNet110                 & 110.00                      & 71.85         \\
ResNet56                  & 56.00                       & 70.57         \\
ResNet44                  & 44.00                       & 70.28         \\
ResNet32                  & 34.00                       & 69.28         \\
ResNet20                  & 20.00                       & 67.14         \\ \hline \hline
\multicolumn{3}{|c|}{Budgeted ResNet}                                   \\ \hline
\multirow{3}{*}{B-ResNet} & 179.00                      & 74.35         \\
                          & 112.00                      & 73.85         \\
                          & 49.00                       & 71.84         \\
                          & 29.00                       & 69.94         \\
                          & 22.00                       & 69.09         \\ \hline
\end{tabular}
\caption{Results for \textit{Distributed computation cost} on CIFAR-100 with $n=2$}
\label{cif100_resnetfab_para2}
\end{table}

\end{document}